\documentclass[10pt,twocolumn,letterpaper]{article}

\usepackage{wacv}              % To produce the CAMERA-READY version
\usepackage{graphicx}
\usepackage{amsmath}
\usepackage{amssymb}
\usepackage{booktabs}

\usepackage{multirow}

\usepackage[pagebackref,breaklinks,colorlinks]{hyperref}

\usepackage[capitalize]{cleveref}
\crefname{section}{Sec.}{Secs.}
\Crefname{section}{Section}{Sections}
\Crefname{table}{Table}{Tables}
\crefname{table}{Tab.}{Tabs.}

\def\wacvPaperID{009} % *** Enter the WACV Paper ID here
\def\confName{WACV}
\def\confYear{2025}

\title{Digi2Real: Bridging the Realism Gap in Synthetic Data Face Recognition via Foundation Models
}

\author{\vspace{3pt} Anjith George and  Sébastien Marcel \\
Idiap Research Institute, Switzerland\\
{\tt\small \{anjith.george,sebastien.marcel\}@idiap.ch}\\
}
\begin{document}
% \maketitle
% \thispagestyle{empty}
% \begin{figure*}[h] % Ensure that the figure takes the full width at the top of the page
%     \centering
%     \includegraphics[width=\linewidth]{figures/DigiReal-Page-2.drawio.png}
%     \caption{Examples of original images from the DigiFace dataset (first row) and the corresponding transformed images using our approach (second row)}
%     \label{fig:synth_samples}
% \end{figure*}

% \begin{figure*}[t!] % Ensure that the figure takes the full width at the top of the page
%     \centering
%     \includegraphics[width=\linewidth]{figures/DigiReal-Page-8.drawio3.png}
%     \caption{The images on the left show an example identity from the DigiFace dataset alongside its realism-enhanced versions, illustrating intra-class variations. On the right, the first row showcases original images from the DigiFace dataset, while the second row presents the corresponding transformed images generated using our approach.
%     }
%     \label{fig:synth_samples}
% \end{figure*}

\maketitle
\begin{figure*}[!htbp] % Ensure that the figure takes the full width at the top of the page
    \centering
    \includegraphics[width=0.99\linewidth]{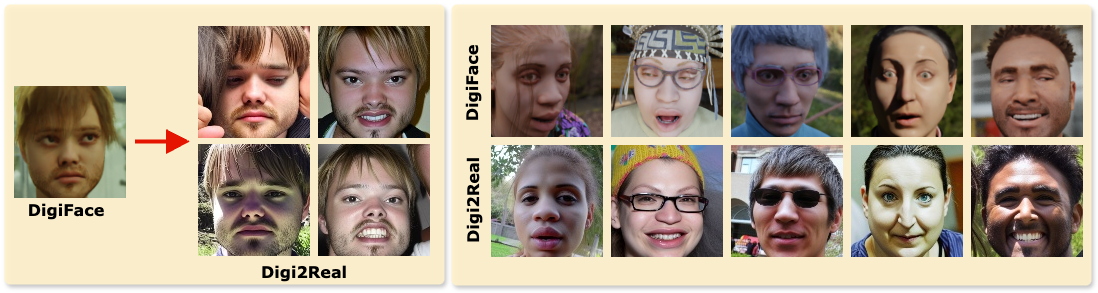}
    \caption{The images on the left show an example identity from the DigiFace dataset alongside its realism-enhanced versions, illustrating intra-class variations. On the right, the first row showcases original images from the DigiFace dataset, while the second row presents the corresponding transformed images generated using our approach.}
    \label{fig:synth_samples}
\end{figure*}
%%%%%%%%% ABSTRACT

\begin{abstract}

The accuracy of face recognition systems has improved significantly in the past few years, thanks to the large amount of data collected and advancements in neural network architectures. However, these large-scale datasets are often collected without explicit consent, raising ethical and privacy concerns. To address this, there have been proposals to use synthetic datasets for training face recognition models. Yet, such models still rely on real data to train the generative models and generally exhibit inferior performance compared to those trained on real datasets. One of these datasets, DigiFace, uses a graphics pipeline to generate different identities and intra-class variations without using real data in model training. However, the performance of this approach is poor on face recognition benchmarks, possibly due to the lack of realism in the images generated by the graphics pipeline. In this work, we introduce a novel framework for realism transfer aimed at enhancing the realism of synthetically generated face images. Our method leverages the large-scale face foundation model, and we adapt the pipeline for realism enhancement. By integrating the controllable aspects of the graphics pipeline with our realism enhancement technique, we generate a large amount of realistic variations—combining the advantages of both approaches. Our empirical evaluations demonstrate that models trained using our enhanced dataset significantly improve the performance of face recognition systems over the baseline. The source code and dataset will be publicly accessible at the following link: \url{https://www.idiap.ch/paper/digi2real}

\end{abstract}

%%%%%%%%% BODY TEXT
\section{Introduction}
\label{sec:intro}

Face recognition (FR) technology has seen widespread adoption due to its accuracy and ease of use. This high performance can be attributed to advancements in deep learning architectures, margin-based loss functions \cite{deng2019arcface,kim2022adaface}, and large-scale public datasets \cite{zhu2021webface260m,guo2016ms}. However, the acquisition of many of these datasets often conducted without explicit user consent, posing significant legal and ethical challenges, particularly in light of the European Union's General Data Protection Regulation (GDPR) and laws in other regions. As a result of these regulations, several datasets have been withdrawn, reducing the availability of training data. Consequently, there is a growing interest in generating high-performance synthetic face recognition datasets for training face recognition models, as evidenced in public competitions on this topic \cite{melzi2024frcsyn, otroshi2024sdfr,deandres2024frcsyn,melzi2024frcsynongoing}.

Over the past few years, several synthetic face datasets have been introduced for training face recognition models. Most of them rely on generative models like StyleGAN \cite{karras2019style} and Diffusion Models \cite{rombach2022high}. These models are often trained on real datasets such as FFHQ \cite{karras2019style} to model the distribution of faces. Most of them also use pretrained face recognition networks in the generation pipeline to induce the notion of identity in the sampling process. However, these datasets suffer from issues like a limited number of identities and limited intra-class variation. Preserving the identity while generating intra-class variations is another challenging issue here as these are sampled from a continuous latent space. These issues result in reduced accuracy of face recognition models trained with these synthetic datasets, which limits the practical use of these models, as the performance is much worse than those trained using real data. It should also be noted that the diversity and accuracy of these models could vary based on the training data used to train the generative model in the pipeline.

DigiFace-1M \cite{bae2023digiface} offers an alternative to generative models for data creation by utilizing a graphics rendering pipeline to produce images without requiring large-scale real images or a face recognition network. This method uses the rendering pipeline described in \cite{wood2021fake} to combine facial geometry, texture, and hairstyle. It facilitates the generation of intra-class variations by rendering additional images with varying poses, expressions, lighting, and accessories. This approach holds significant potential for creating a large number of identities with diverse intra-class variations and ethnicities. Interestingly, it also allows for controlled generation processes by selecting specific attributes. However, the primary limitation of this approach is the lack of realism in the generated samples, resulting in poor face recognition performance for models trained with this dataset. The computational requirement to produce these samples is another constraint. 

In our proposed method, we introduce a novel approach to enhance the realism of the procedurally generated DigiFace dataset. By reusing existing DigiFace samples as our source dataset, we eliminate the need for recomputation and demonstrate through empirical evidence that our approach substantially boosts performance. Our approach significantly improves face recognition performance compared to DigiFace and achieves comparable performance to other state-of-the-art synthetic data methods. 

The main contributions of this work are listed below:

\begin{itemize}

    \item We introduce a novel framework for generating realistic sample images from those generated by a graphics pipeline.
    \item We introduce a new synthetic face recognition dataset called Digi2Real with a large number of identities and intra-class variations
    \item We conduct a thorough analysis using the generated dataset, demonstrating the advantages of our approach.
\end{itemize}

Additionally, the dataset and the code necessary to reproduce the results will be made publicly available \footnote{ \url{https://www.idiap.ch/en/scientific-research/data/digi2real}}.

% \begin{figure}[!h] 
%                \centering
%                \includegraphics[width=0.99\columnwidth]{figures/DigiReal-Page-4.drawio.pdf}
%                \caption{The images on the left show variations of the one identity from the DigiFace dataset, and the images on the right show the realism-enhanced images generated from our approach}
%                \label{fig:synth_variations} 
% \end{figure}

\section{Related works}

Recent studies have extensively explored the generation of synthetic datasets as a solution to the legal and ethical constraints associated with the use of real data. Many of these approaches employ StyleGAN or Diffusion models within their generation pipelines. In this section, we provide a brief review of the synthetic face datasets.

 SynFace \cite{qiu2021synface} focused on developing facial recognition (FR) models using synthetic data. It evaluated DiscoFaceGAN \cite{deng2020disentangled} to analyze intra-class variance and the domain gap between real and synthetic images. SynFace enhanced DiscoFaceGAN by introducing identity and domain mixups, techniques that blend image features to create new synthetic faces. SFace \cite{boutros2022sface} introduced a class-conditional synthetic GAN designed for generating class-labeled synthetic images. By leveraging synthetic data, the authors trained supervised FR models that demonstrated reasonable performance. In \cite{boutros2023idiff}, the authors introduced IDiff-Face, which leverages conditional latent diffusion models for generating synthetic identities. The approach uses a two-stage pipeline: first, an autoencoder is trained, and then its latent space is used to train the diffusion model. The generation process is conditioned by projecting training images into a low-dimensional space and incorporating them into the intermediate stages of the diffusion models via a cross-attention mechanism. A dropout mechanism is employed to prevent overfitting to identity-specific features. In \cite{melzi2023gandiffface}, the authors presented the GANDiffFace dataset, which integrates StyleGAN and diffusion-based methods. The StyleGAN model is initially used to create realistic identities with specified demographic distributions. These generated images are then used to fine-tune diffusion models, synthesizing various identities. The latent space is manipulated to sample identities across different demographic groups. ExFaceGAN \cite{boutros2023exfacegan} proposes a novel method for disentangling identity within the latent space of a StyleGAN, allowing for the synthesis of multiple variations. This technique involves learning identity decision boundaries by splitting the latent space into two distinct subspaces. Images generated from either side of the boundary can represent the same or different identities without the need for separate classifiers. These subspaces represent image transformations that maintain the original identity, making the approach compatible with any GAN-based generator. IDNet \cite{kolf2023identity} introduces a three-player generative adversarial network (GAN) framework that incorporates identity information into StyleGAN's generation process. Besides learning the distribution of real data, a third player—a face recognition network pre-trained on the CASIA-WebFace dataset—ensures the generator produces identity-separable images. In this setup, only the classification layers of the face recognition network are trained, enhancing the generator's ability to create distinct identities.  Works such as Synthdistill \cite{shahreza2023synthdistill,shahreza2024knowledge}, utilized the generation of synthetic data using the feedback mechanism in a loop to train lightweight models by generating challenging samples. In \cite{geissbuhler2024synthetic}, the authors introduce a novel method for sampling the latent space of a StyleGAN, inspired by the physical motion of soft particles under stochastic Brownian forces. This framework allows for the inclusion of multiple constraints in the sampling process. They propose an algorithm named DisCo, which combines identity Dispersion with latent directions augmentation to synthesize both diverse identities and intra-class variations. The identity Dispersion algorithm samples the latent space around each class's identity reference latent vector, optimizing these intra-class samples to be close in the embedding space. Their method achieved accuracy comparable to state-of-the-art methods for synthetic face recognition using StyleGAN.

One major limitation of generative approaches is the difficulty in synthesizing a large number of unique identities and producing identity-consistent intra-class variations. In contrast to these generative approaches, DigiFace \cite{bae2023digiface} proposes a rendering-based pipeline to generate synthetic face identities. DigiFace used 511 3D face scans, obtained with consent, to construct a parametric model of facial geometry and texture. From these scans, a parametric generative face model was created, capable of producing random 3D faces combined with artist-designed textures, hair, and accessories, and rendered in various environments. Distinct combinations of facial features, eye color, and hairstyles define unique identities, effectively reducing diversity and label noise. This pipeline generated a dataset of 1.22 million images featuring 110,000 unique identities. Each identity was created using random combinations of facial geometry, texture, hairstyle, and other variations, rendered in diverse environments. Although this rendering-based approach is computationally intensive, it offers greater flexibility by enabling the creation of a large number of identities and identity-consistent variations, with user-defined sampling of pose, expression, demographics, and more. However, the generated samples lack realism, often resulting in a domain gap and lower performance when training face recognition models.

In the work by Rahimi et al. \cite{rahimi2024synthetic}, image-to-image translation techniques were used to enhance the realism of images from the DigiFace dataset. They aimed to enhance DigiFace images using a pretrained image enhancement network, without relying on a pretrained face recognition network or real face data with identity labels. Models like CodeFormer, designed for image enhancement, were utilized in their pipeline. Although there was a performance improvement, the results remain significantly below the performance of models trained with real data, limiting the practical application of this approach.
 
From the above discussion, it is evident that using a graphics rendering pipeline is a powerful method for generating identity-consistent and diverse large-scale datasets. However, this approach is computationally expensive and lacks realism. To address these issues, we propose a new method that reuses DigiFace images and reduces the realism gap. Our approach eliminates the need for re-rendering images, combining the strengths of both rendering and generative models to achieve better results.

\begin{figure*}[!h] 
               \centering
               \includegraphics[width=0.99\textwidth]{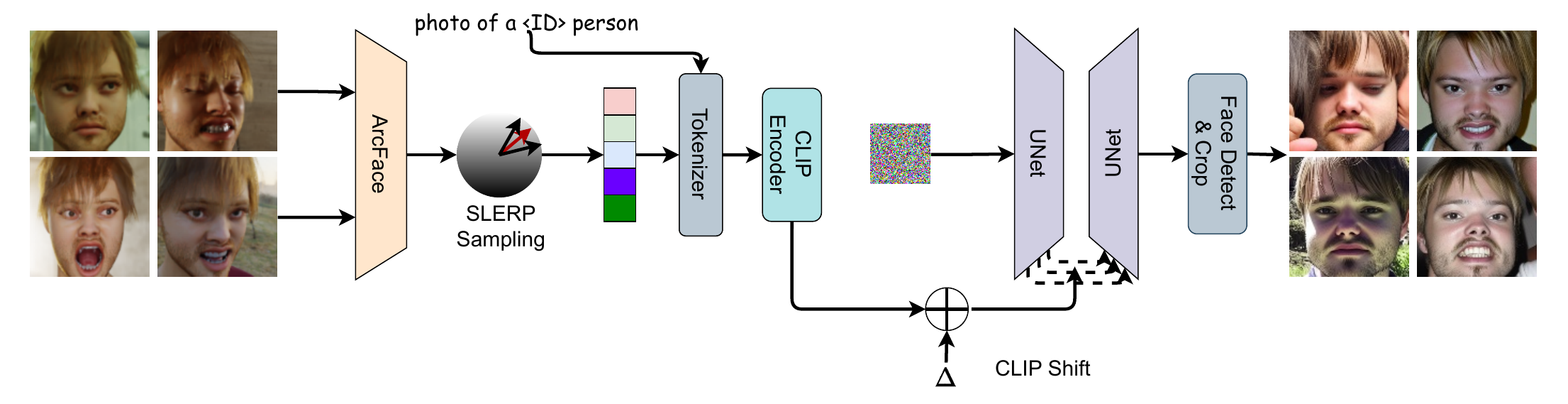}
               \caption{Different stages of the proposed generation pipeline: We start with the original images from DigiFace and generate a class prototype and intra-class variations. A pre-trained Arc2Face model for generating identity-conditioned images, we add the CLIP shift to enhance realism. }
               \label{fig:framework} 
\end{figure*}

\section{Proposed method}

The key insight of our approach is to reuse procedurally generated identities from a graphics pipeline and enhance their realism to reduce the domain gap. The stages of our pipeline are illustrated in Fig. \ref{fig:framework}. DigiFace1M provides an elaborate pipeline for generating synthetic identities and their variations, allowing us to obtain a large number of identities from this dataset. Additionally, we generate variations by interpolating between multiple images of an identity within the embedding space. Using a pre-trained foundation model, specifically the Arc2Face \cite{papantoniou2024arc2face} model, we synthesize identity-consistent images from these interpolated embeddings. We further enhance the realism of the generated images by modifying the intermediate CLIP space. The resulting dataset consists of various variations suitable for training a face recognition model. Detailed descriptions of the different components are presented in this section.

\textbf{Sampling Identities} To generate samples with multiple identities, we draw from the identities in the DigiFace1M dataset. This ensures that the identities are unique and come from a purely synthetic dataset, enabling the creation of a large set of identities. It’s worth noting that while we reuse DigiFace, this framework can be extended to generate a vast number of identities with various variations in pose, ethnicity, gender, and more. This approach offers significantly more flexibility compared to directly sampling from the latent space of generative models. To generate realistic face images from the sampled images, we need to transform them to preserve the identity while introducing enough intra-class variations.

\textbf{Identity Consistent Face Generation} The objective of this stage is to generate images that share the same identity as the original images from the DigiFace dataset. Arc2Face \cite{papantoniou2024arc2face} introduces an identity-conditioned face foundation model that can generate photorealistic images based on face embeddings from an ArcFace model. They leverage Stable Diffusion \cite{rombach2022high}, a text-conditioned latent diffusion model that utilizes text embeddings from a CLIP encoder \cite{radford2021learning} to generate photorealistic images. Stable Diffusion employs a UNet \cite{ronneberger2015u} as its backbone and incorporates cross-attention layers to map auxiliary features from the CLIP embeddings into its intermediate layers for conditional generation. The Arc2Face model uses a simple prompt, \textit{photo of a \textless id\textgreater person} after tokenization. The ``\textless id\textgreater'' token embedding is replaced with the ArcFace embedding vector to obtain the token embeddings. These embeddings are then fed into the CLIP encoders to obtain the CLIP latent representation.

This process effectively transforms ArcFace embeddings into the CLIP latent space. The Arc2Face model was trained on a large-scale, upscaled version of WebFace42M \cite{zhu2021webface260m} and fine-tuned on FFHQ and CelebA-HQ. Despite being trained on a substantial dataset, their ablation studies demonstrate that their approach does not memorize the training data. The CLIP latents are used with the UNet to generate photorealistic images.

\begin{figure}[t!]%[htp!]
\begin{center}
               \includegraphics[width=0.4\textwidth]{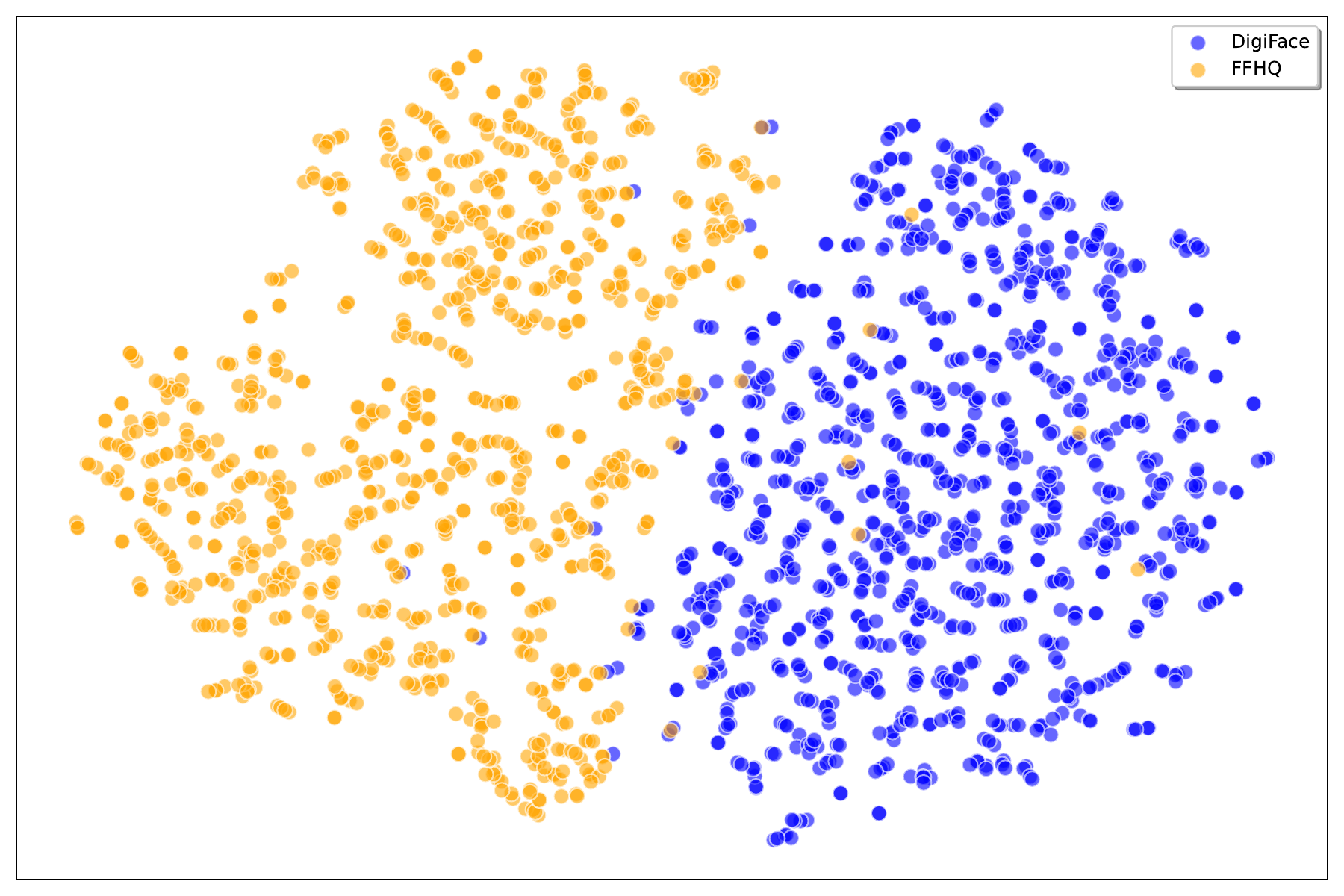}
               \caption{T-SNE plots of samples from the FFHQ and DigiFace datasets in the CLIP latent space show a clear difference in the distribution of embeddings between the two datasets. }
               \label{fig:tsneclip} 
\end{center}
\end{figure}

\textbf{Reducing Realism Gap} Despite the enhanced realism in projecting and reconstructing synthetic identities, they still lacked the visual quality of natural images. In Fig. \ref{fig:tsneclip}, we plot the T-SNE of CLIP representations from the encoder for 1000 images sampled from both FFHQ and DigiFace. As seen in the figure, there’s a disparity between the distributions of embeddings for real images and those from DigiFace. To address this gap, we propose calculating the offset in the CLIP space and estimating a shift vector. This approach is similar to the Domain Gap Embeddings proposed in \cite{doge2024} (\(\Delta\)).

\begin{equation}
    \Delta = \frac{1}{N} \sum_{i=1}^{N} \left( \text{CLIP}_{\text{FFHQ}_i} \right) - \frac{1}{N} \sum_{i=1}^{N} \left( \text{CLIP}_{\text{DigiFace}_i} \right)
\end{equation}

The estimated shift values, $\Delta$, are added to correct the distribution in our generation pipeline as follows:

\begin{equation}
    CLIP_{corrected} = CLIP_{DigiFace} + \Delta
\end{equation}

\begin{figure}[t!]%[htp!]
\begin{center}
               \includegraphics[width=0.99\columnwidth]{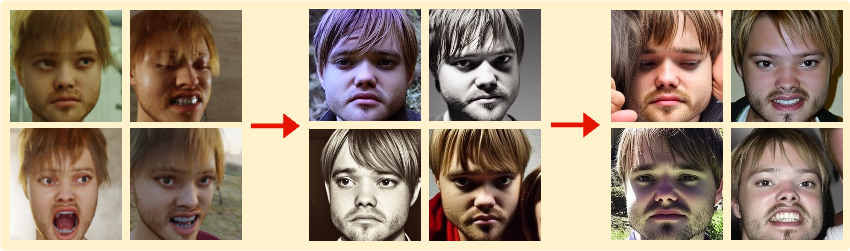}
               \caption{The set of images on the left shows images from DigiFace, the middle set shows images generated using identity-preserving sampling, and the right set shows images generated with CLIP shift added.}
               \label{fig:stages} 
\end{center}
\end{figure}

Figure \ref{fig:stages} shows the improvement in realism with the additional correction term in the CLIP space.

\textbf{Generating Intra-class Variations} Challenging intra-class variations is an important aspect that is needed to train robust face recognition networks. To generate intra-class variations, we use the intra-class samples from the DigiFace dataset. First, we generate the ArcFace embeddings for all samples of the same subject and normalize them to the unit hypersphere. One naive way would be to generate samples from each of the images present. However, we have observed that sampling from different images as starting points resulted in identity drift (especially for images in extreme off-pose and extreme expressions). Hence, we opted to introduce a class prototype vector for each identity computed as the mean vector embedding space and introduce variability by sampling along the geodesic on the unit hypersphere to generate variations. Specifically, we perform spherical interpolation (SLERP) \cite{shoemake1985animating} on the unit sphere using the following equation:

Let $\mu$ be the class mean on the unit hypersphere, and \(e_i, e_i \in \{1, 2, \ldots, K\}\) represent embeddings corresponding to samples in DigiFace. We compute intra-class variation by sampling along the directions of samples present in the dataset. For an interpolation factor \(\lambda\), the interpolation equation is given by:

\begin{equation}
    \mathbf{v}(\lambda) = \frac{\sin((1 - \lambda) \theta_i)}{\sin(\theta_i)} \mu + \frac{\sin(\lambda \theta_i)}{\sin(\theta_i)} e_i
\end{equation}

\noindent where $\theta_i$ is the angle between $\mu$ and $e_i$, defined as:

\begin{equation}
    \theta_i = \cos^{-1}(\mu \cdot e_i)
\end{equation}

\begin{figure}[t!]%[htp!]
\begin{center}
               \includegraphics[width=0.79\columnwidth]{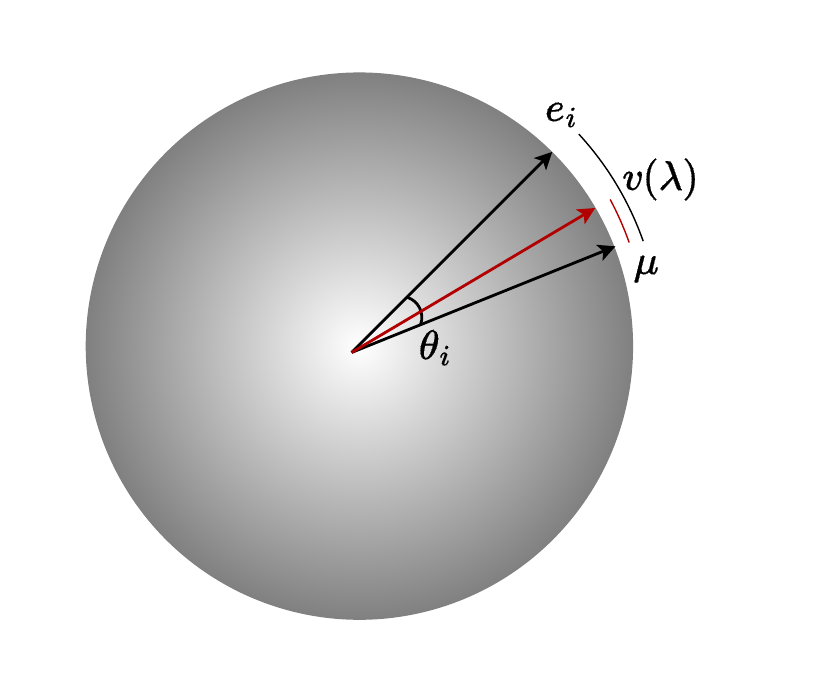}
               \caption{Intra-identity sampling is performed using spherical linear interpolation on the unit sphere, from the class prototype to the direction of the image samples in DigiFace.}
               \label{fig:slerp} 
\end{center}
\end{figure}

\noindent The value of $\lambda$ is sampled from a Beta distribution $\beta(\alpha, \beta))$

\begin{equation}
    \lambda \sim \text{Beta}(\alpha, \beta)
\end{equation}
for value of $\alpha=\beta=1$, it is equivalent to uniform distribution in the range $[0,1]$.

\textbf{Dataset Generation} For each subject in the DigiFace dataset, we select five images and project them into the ArcFace embedding space. We perform SLERP (Spherical Linear Interpolation) sampling based on the Beta distribution to generate intra-class variations. The resulting sampled latents are fed into the CLIP encoder to produce a latent representation. This representation is then modified using the estimated \(\Delta\) parameter to generate five images per latent. These images undergo post-processing, where faces are detected, cropped, and aligned using MTCNN to a uniform resolution of $112 \times 112$. We process up to 87K identities from the original DigiFace dataset.

\section{Experiments}

We trained face recognition models using the dataset generated through our proposed approach. This section provides a detailed evaluation, including ablation studies and comparisons with state-of-the-art methods.

\textbf{Evaluation Datasets} We evaluate the performance of the model using several benchmarking datasets, including Labeled Faces in the Wild (LFW) \cite{huang2008labeled}, Cross-age LFW (CA-LFW) \cite{zheng2017cross}, CrossPose LFW (CP-LFW) \cite{zheng2018cross}, Celebrities in Frontal-Profile in the Wild (CFP-FP) \cite{sengupta2016frontal}, AgeDB-30 \cite{moschoglou2017agedb}, IARPA Janus Benchmark-B (IJB-B) \cite{whitelam2017iarpa}, and IARPA Janus Benchmark-C (IJB-C) \cite{maze2018iarpa}. 

Following previous works, we report accuracies for datasets such as LFW, CA-LFW, CP-LFW, CFP-FP, and AgeDB-30. For the IJB-B and IJB-C datasets, we report the True Accept Rate (TAR) at a False Accept Rate (FAR) of $1 \times 10^{-4}$.

\textbf{Training details} In all experiments, we adopt the IResNet50 architecture to ensure comparability with previous literature. Specifically, we employ AdaFace \cite{kim2022adaface} as the loss function. The models were trained for 30 epochs with a batch size of 256, following a cosine learning rate schedule. For the loss, we used the default hyperparameter settings as specified in the AdaFace \cite{kim2022adaface} paper for our experiments. Models were trained using NVIDIA RTX 3090 GPUs.

\subsection{Ablation Experiments}

A wide range of design choices are involved in generating the final dataset, so we performed a series of experiments to determine the optimal parameter set. The specifics are outlined in this subsection.

\begin{table}[!htb]
\centering
\caption{Effect of varying the number of images per identity, experiment was performed with 10K identities.}
%All with 10 K Ids, change in images per id, alpha =1
\resizebox{0.99\columnwidth}{!}{

\begin{tabular}{lrrrrrrr}
\toprule
\textbf{Images per ID} & \textbf{LFW} & \textbf{CALFW} & \textbf{CPLFW} & \textbf{AgeDB30} & \textbf{CFP-FP} & \textbf{IJB-B} & \textbf{IJB-C} \\
\midrule
10 &  97.40 &  84.80 &  78.40 &    80.63 &   81.94 &  66.66 &  70.95 \\
20 &  97.95 &  85.97 &  79.25 &    81.83 &   84.66 &  67.86 &  72.28 \\
25 &  98.03 &  85.80 &  79.17 &    81.70 &   84.94 &  30.81 &  43.80 \\
\bottomrule
\end{tabular}
}
\label{tab:verification_num_image_selection}
\end{table}

\textbf{Number of Images per Subject} Given that the smallest sample count per subject in DigiFace was five, we performed an experiment to determine the optimal number of images per subject. We generated 10,000 identities with 10, 20, and 25 images each, then evaluated face recognition performance. The results, shown in Table \ref{tab:verification_num_image_selection}, indicate that while performance remains stable in high-quality datasets, IJB-C and IJB-B performance declines with higher image counts. Based on these results, we selected 20 as the optimal number of samples for further experiments.

\begin{table}[!htb]
\centering

\caption{Effect of changing the value of $\alpha$, experiment performed with 10K identities.}
\resizebox{0.99\columnwidth}{!}{

\begin{tabular}{lrrrrrrr}
\toprule
\textbf{$\alpha$} & \textbf{LFW} & \textbf{CALFW} & \textbf{CPLFW} & \textbf{AgeDB30} & \textbf{CFP-FP} & \textbf{IJB-B} & \textbf{IJB-C} \\
\midrule
1 &  97.40 &  84.80 &  78.40 &    80.63 &   81.94 &  66.66 &  70.95 \\
2 &  97.87 &  86.18 &  78.78 &    82.20 &   85.10 &  68.87 &  73.09 \\
\bottomrule
\end{tabular}
}
\label{tab:verification_alpha}
\end{table}

\textbf{ Value of $\alpha$} The hyperparameter $\alpha$ controls the distribution of samples around each class prototype, with samples selected in the direction of DigiFace embeddings and drawn from a beta distribution parameterized by $\alpha$. We performed an experiment with 10,000 identities, each having 20 images, using $\alpha$ values of 1 and 2. The results, shown in Table \ref{tab:verification_alpha}, indicate that an $\alpha$ of 2 yielded improved performance, particularly on IJB-B and IJB-C, so we chose $\alpha = 2$ for subsequent experiments.

\begin{table}[!htb]
\centering

\caption{Impact of varying number of identities generated on the face recognition performance.}
\resizebox{0.99\columnwidth}{!}{

\begin{tabular}{lrrrrrrr}
\toprule
\textbf{No of ID's} & \textbf{LFW} & \textbf{CALFW} & \textbf{CPLFW} & \textbf{AgeDB30} & \textbf{CFP-FP} & \textbf{IJB-B} & \textbf{IJB-C} \\
\midrule
10K                        &  97.87 &  86.18 &  78.78 &    82.20 &   85.10 &  68.87 &  73.09 \\
20K                        &  98.12 &  86.60 &  81.17 &    82.77 &   87.20 &  57.15 &  63.86 \\
30K                         &  98.17 &  88.10 &  81.68 &    84.63 &   87.84 &  53.20 &  59.86 \\
40K                         &  98.52 &  88.25 &  82.62 &    84.65 &   88.63 &  50.02 &  57.30 \\
50K                         &  98.43 &  88.58 &  82.25 &    84.60 &   88.86 &  32.38 &  37.17 \\
60K                         &  98.28 &  88.27 &  83.58 &    85.73 &   88.76 &  43.92 &  48.95 \\
70K                         &  97.83 &  88.02 &  82.13 &    84.92 &   88.96 &  36.47 &  40.66 \\
80K                         &  97.78 &  88.68 &  81.87 &    85.78 &   89.56 &  31.01 &  34.15 \\
87K                         &  98.17 &  88.78 &  82.85 &    86.05 &   89.53 &  37.52 &  42.38 \\

\bottomrule
\end{tabular}
}
\label{tab:verification_ids}
\end{table}

\begin{table}[!htb]
\centering

\caption{Face recognition performance with the sorted list of identities.}
\resizebox{0.99\columnwidth}{!}{

\begin{tabular}{lrrrrrrr}
\toprule
\textbf{No of ID's} & \textbf{LFW} & \textbf{CALFW} & \textbf{CPLFW} & \textbf{AgeDB30} & \textbf{CFP-FP} & \textbf{IJB-B} & \textbf{IJB-C} \\
\midrule
10K           &  98.12 &  86.32 &  80.77 &    80.98 &   84.87 &  66.92 &  72.79 \\
20K           &  98.58 &  87.22 &  82.65 &    84.72 &   87.49 &  70.14 &  75.80 \\
30K           &  98.37 &  88.87 &  83.95 &    84.88 &   88.83 &  62.94 &  67.82 \\
40K           &  98.77 &  88.18 &  83.90 &    85.13 &   88.60 &  44.21 &  52.39 \\
50K           &  98.50 &  88.62 &  84.10 &    85.32 &   89.21 &  37.52 &  43.75 \\
60K           &  98.47 &  88.77 &  83.90 &    85.13 &   89.57 &  38.53 &  42.40 \\
70K           &  98.12 &  88.58 &  83.02 &    85.68 &   89.34 &  31.26 &  33.91 \\
80K           &  97.57 &  88.17 &  82.50 &    85.77 &   89.77 &  33.55 &  36.10 \\
% LAST10K       &  97.52 &  85.47 &  76.40 &    82.53 &   83.73 &  53.12 &  60.77 \\
\bottomrule
\end{tabular}
}
\label{tab:verification_ids_sorted}
\end{table}

\textbf{Number of identities} Another factor in synthetic dataset generation is the number of identities created. Since our approach relies on the identities in the DigiFace dataset, the total number of generated identities is limited. We conducted an experiment to evaluate the impact of generating different identity sets, with results shown in Table \ref{tab:verification_ids}. While the DigiFace dataset had 110K identities, we capped the number of identities at a maximum of 87K, as adding more led to diminishing returns, and performance on the IJB-B and IJB-C datasets decreased with additional identities. 

\begin{figure}[t!]%[htp!]
\begin{center}
               \includegraphics[width=0.47\textwidth]{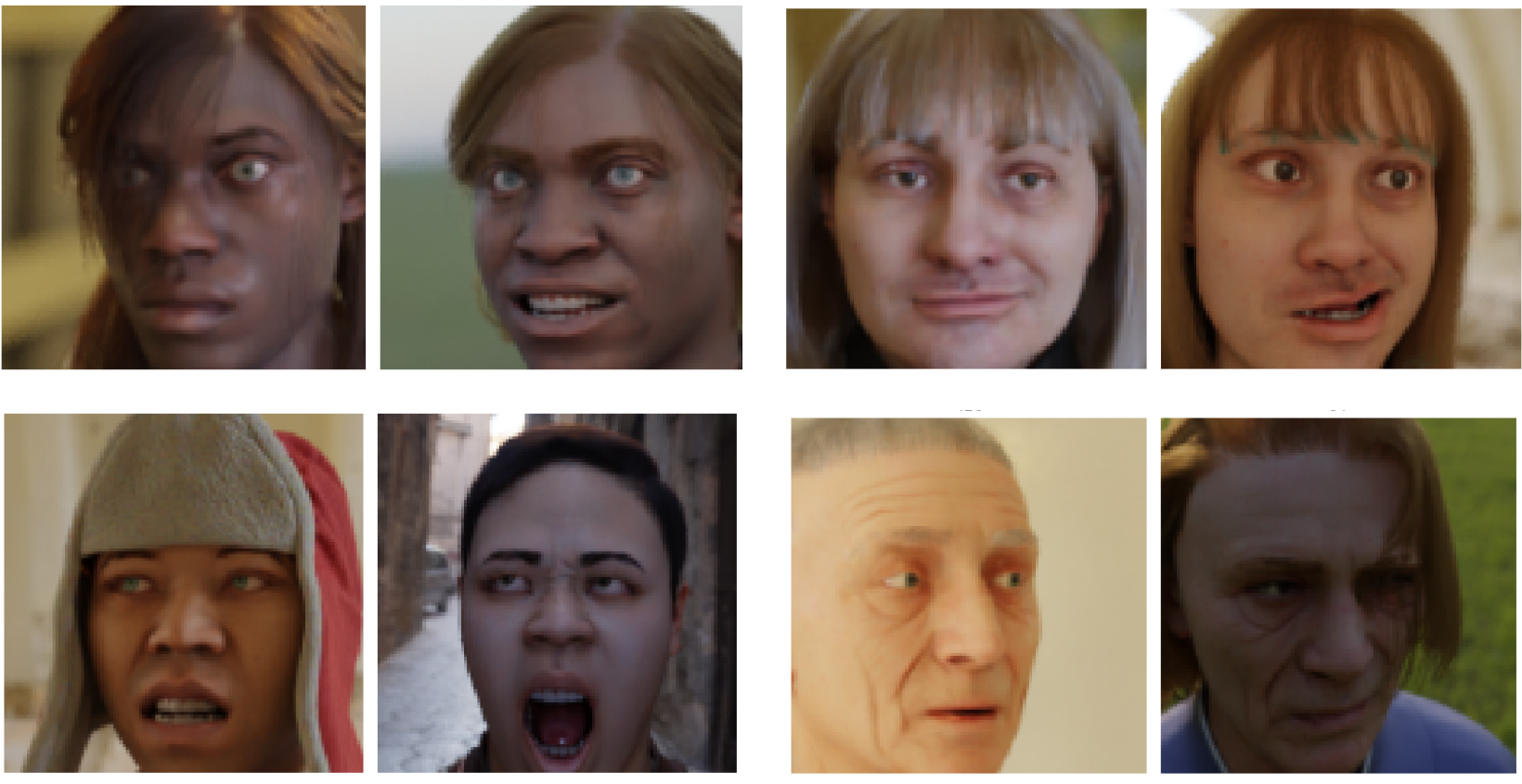}
               \caption{Examples of images from DigiFace the are labeled with distinct identities, but the only differences are attributes like skin tone and accessories.}
               \label{fig:digisimilar} 
\end{center}
\end{figure}

The performance degradation observed with an increased number of identities was counterintuitive, so we investigated this issue further by examining the generated samples. Our analysis revealed that some identities in the original DigiFace dataset were highly similar. As DigiFace generates data through random combinations of attributes, it sometimes creates identities that are visually alike but are classified as different, such as faces with similar shapes but varying skin tones. Some examples are shown in Fig. \ref{fig:digisimilar}. This overlap complicates the training of robust face recognition models. 

To address this, we filtered the dataset based on identity similarity. We sorted identities based on their cosine similarities with other identities and created a set of identities from this sorted list for our experiments. Using this filtered set, we repeated our experiments and observed performance improvements in the IJB-B and IJB-C datasets, as shown in Table \ref{tab:verification_ids_sorted}. In this case, 20,000 samples proved optimal. This sorted set of identities was used for all subsequent comparisons, denoted as Digi2Real-20K. % Some examples of these are shown in Fig. XX. % TODO

\subsubsection{Evaluations}

\begin{figure*}[!htb]
    \centering
    \begin{subfigure}{0.49\textwidth}
        \centering
        \includegraphics[width=\linewidth]{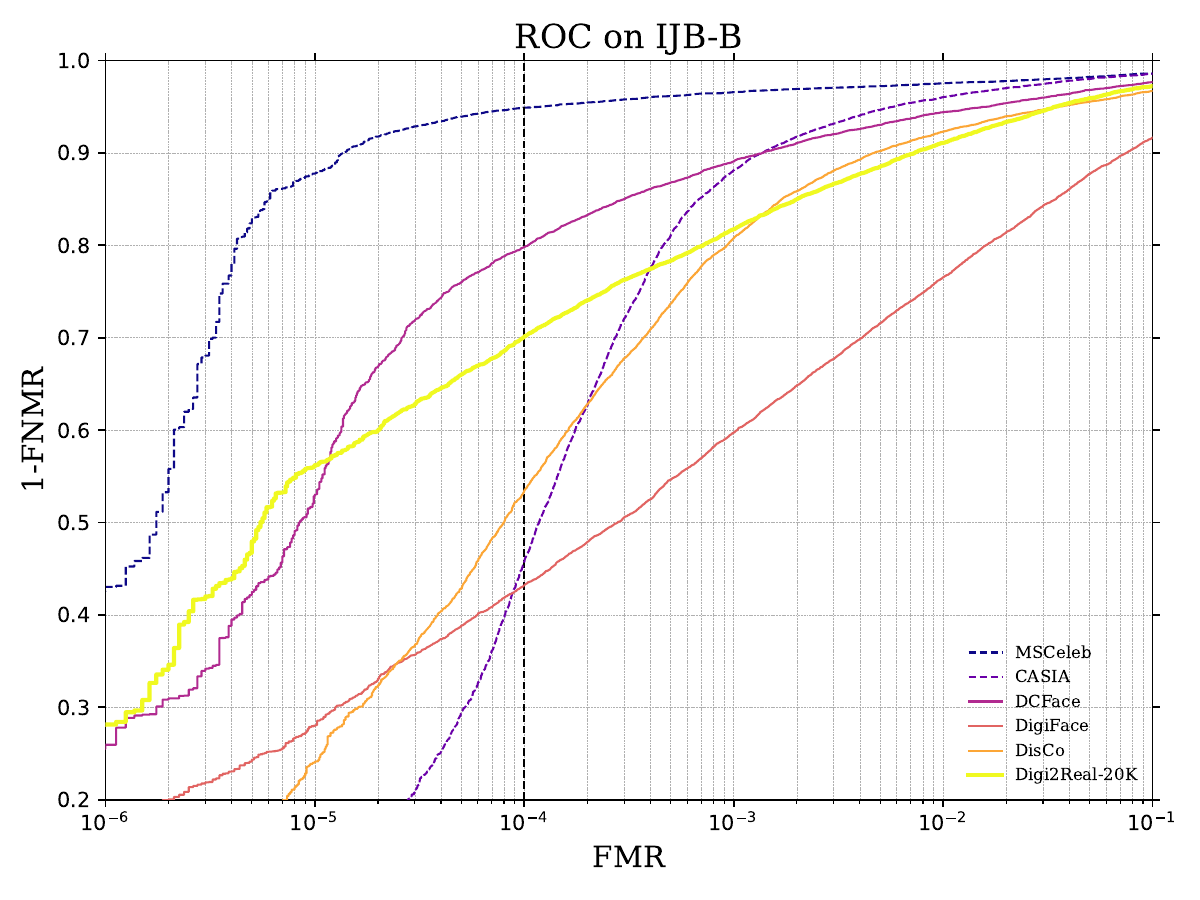}
        \caption{IJB-B}
        \label{fig:ijbb}
    \end{subfigure}\hfill
    \begin{subfigure}{0.49\textwidth}
        \centering
        \includegraphics[width=\linewidth]{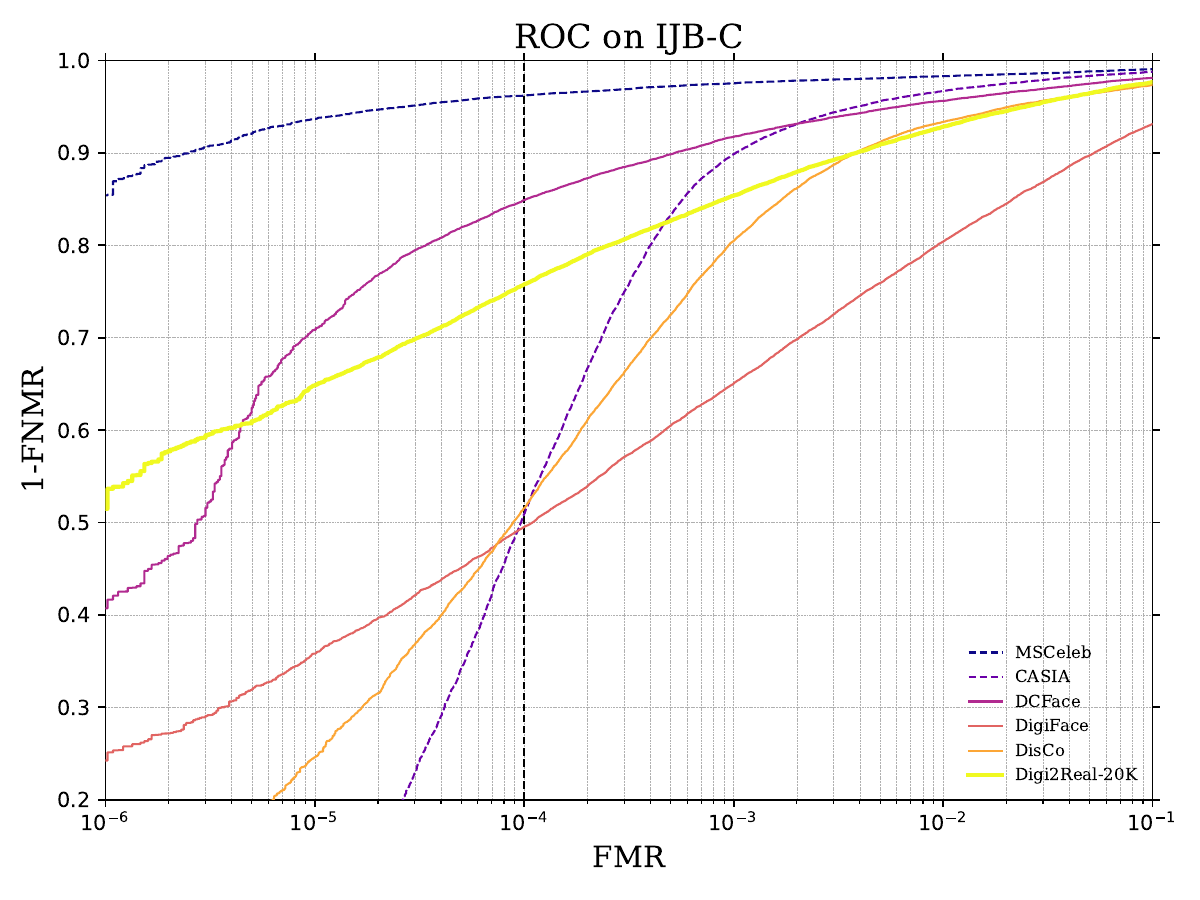}
        \caption{IJB-C}
        \label{fig:ijbc}
    \end{subfigure}
    \caption{Comparison of IJB-B and IJB-C performance}
    \label{fig:ijbb-ijbc-comparison}
\end{figure*}

In this section, we compare the performance of our generated Digi2Real-20K dataset with other synthetic and real datasets on standard benchmarks. Specifically, Digi2Real-20K is the enhanced version of DigiFace, generated using our proposed realism-enhancement approach, consisting of 20K identities with 20 images per identity.

\begin{table}[!htb]
\centering
\caption{Performance improvement from DigiFace to Digi2Real-20K}
\resizebox{0.99\columnwidth}{!}{
\begin{tabular}{lrrrrrrr}
\toprule
\textbf{Dataset} & \textbf{LFW} & \textbf{CALFW} & \textbf{CPLFW} & \textbf{AgeDB30} & \textbf{CFP-FP} & \textbf{IJB-B} & \textbf{IJB-C} \\
\midrule

DigiFace1M \cite{bae2023digiface}               &  91.93 &  75.80 &  73.13 &    71.12 &   78.89 &  43.23 &  49.53 \\
Digi2Real-20K           &  98.58 &  87.22 &  82.65 &    84.72 &   87.49 &  70.14 &  75.80 \\

\bottomrule
\end{tabular}
}
\label{tab:verification_sota_digiface}
\end{table}

\textbf{Comparison with DigiFace} Since Digi2Real-20K was created by enhancing the realism of DigiFace, it is essential to evaluate its face recognition performance against DigiFace. The results of this comparison are presented in Table \ref{tab:verification_sota_digiface}. Our realism transfer approach shows a substantial performance improvement, even though only 20,000 identities from DigiFace were used. This indicates that enhancing realism considerably boosts face recognition performance.

\begin{table*}[!htb]
    \centering
    \caption{Comparison with State-of-the-Art synthetic and real datasets (SOTA)}
    \resizebox{0.79\textwidth}{!}{

    \begin{tabular}{llrrrrrrr}
        \toprule
        &\textbf{Dataset} & \textbf{LFW} & \textbf{CALFW} & \textbf{CPLFW} & \textbf{AgeDB30} & \textbf{CFP-FP} & \textbf{IJB-B (E-4)} & \textbf{IJB-C (E-4)} \\
        \midrule
        \multirow{2}{*}{\rotatebox{90}{\textbf{Real}}} & MSCeleb \cite{guo2016ms}         &  99.77 &  96.05 &  92.13 &    97.78 &   95.74 &  94.91 &  96.20 \\
        & CASIA WebFace \cite{yi2014learning}   &  99.47 &  93.83 &  90.07 &    94.40 &   94.99 &  45.80 &  50.94 \\
        \midrule
        \multirow{9}{*}{\rotatebox{90}{\textbf{Synthetic}}} & DCFace \cite{kim2023dcface}                                      &  98.97 &  92.37 &  84.40 &    91.03 &   88.03 &  79.82 &  84.91 \\
& GANDiffFace \cite{melzi2023gandiffface}                                 &  93.52 &  76.95 &  74.33 &    68.32 &   76.91 &  47.85 &  52.95 \\
& DisCo \cite{geissbuhler2024synthetic}                                      &  98.83 &  93.38 &  81.52 &    92.63 &   83.39 &  53.45 &  51.55 \\
& IDiffFace \cite{boutros2023idiff}                                   &  97.72 &  89.70 &  80.77 &    84.10 &   81.19 &  65.08 &  67.11 \\
& ExFaceGAN \cite{boutros2023exfacegan}                                   &  84.65 &  68.42 &  65.27 &    56.22 &   65.06 &  24.79 &  26.61 \\
& DigiFace1M \cite{bae2023digiface}                                    &  91.93 &  75.80 &  73.13 &    71.12 &   78.89 &  43.23 &  49.53 \\
& Syn2Auth-DF-CF \cite{rahimi2024synthetic}                             &  93.18 &  77.43 &  76.82 &    75.53 &   82.33 &  54.30 &  59.96 \\
& SFace \cite{boutros2022sface}                                       &  93.35 &  77.20 &  74.47 &    70.45 &   76.97 &  18.79 &  12.24 \\ 
& \textbf{Digi2Real-20K (ours)}           &  98.58 &  87.22 &  82.65 &    84.72 &   87.49 &  70.14 &  75.80 \\

        \bottomrule
    \end{tabular}
    }
    \label{tab:verification_sota}
\end{table*}

\textbf{Comparison with State of the Art} We evaluate the performance of face recognition networks trained on various synthetic datasets, including our proposed dataset. Specifically, we compare results against other synthetic datasets generated using StyleGAN and Diffusion-based methods, and we also provide context by including performance on real datasets. The results are presented in Table \ref{tab:verification_sota}. Our proposed dataset demonstrates superior performance over many alternatives, with particularly notable improvements on the IJB-B and IJB-C benchmarks, where it ranks second only to the DCFace dataset. While performance on high-quality datasets like LFW remains comparable to other methods, our approach significantly outperforms many synthetic datasets on IJB-B and IJB-C, achieving verification rates of 70.14\% and 75.80\%, respectively. The ROC plots for comparison with other synthetic and real datasets are shown in Fig. \ref{fig:ijbb-ijbc-comparison}.

\begin{table}[!htb]
\centering
\caption{Comparison with state of the art methods on RFW dataset}

\resizebox{0.99\columnwidth}{!}{
\begin{tabular}{lrrrr|rr}
\toprule
\textbf{Dataset} &  \textbf{African} &  \textbf{Asian} &  \textbf{Caucasian} &  \textbf{Indian} &     \textbf{Mean} &       \textbf{Std.} \\
\midrule
MSCeleb \cite{guo2016ms}                                   &         98.32 &       97.85 &           99.12 &        97.90 &  98.29 &  0.51 \\
CASIA WebFace \cite{yi2014learning}                                        &         88.35 &       86.32 &           94.20 &        89.12 &  89.49 &  2.90 \\ \hline

DCFace \cite{kim2023dcface}                                      &         80.73 &       84.03 &           89.50 &        87.20 &  \textbf{85.37} &  3.31 \\
GANDiffFace \cite{melzi2023gandiffface}                                &         64.60 &       70.72 &           75.38 &        71.55 &  70.56 &  3.87 \\
DisCo \cite{geissbuhler2024synthetic}                                      &         79.53 &       83.28 &           88.65 &        84.83 &  84.07 &  3.27 \\
IDiffFace \cite{boutros2023idiff}                                   &         75.23 &       79.87 &           85.20 &        80.68 &  80.25 &  3.54 \\
ExFaceGAN  \cite{boutros2023exfacegan}                                 &         55.63 &       63.82 &           64.62 &        62.18 &  61.56 &  3.54 \\
DigiFace1M \cite{bae2023digiface}                                   &         65.55 &       69.60 &           72.58 &        70.00 &  69.43 &  2.52 \\

Syn2Auth-DF-CF \cite{rahimi2024synthetic}                            &         69.87 &       73.27 &           76.90 &        73.52 &  73.39 &  2.49 \\
SFace \cite{boutros2022sface}                                       &         64.60 &       69.68 &           74.32 &        67.87 &  69.12 &  3.51 \\ 

\textbf{Digi2Real-20K (ours)}  & 80.67 & 79.95 & 85.33 & 81.40 & 81.84 & 2.08 \\

\bottomrule
\end{tabular}
}

\label{tab:verification_rfw}
\end{table}

\textbf{Performance on RFW} We also assess recognition bias across different demographic groups using the Racial Faces in-the-Wild (RFW) dataset \cite{Wang_2019_ICCV}. We measured recognition performance for each group, calculating verification accuracies for Asian, African, Caucasian, and Indian identities, along with the average accuracy and standard deviation. As shown in Table \ref{tab:verification_rfw}, while a performance gap remains compared to models trained on real data, our approach reduces the disparity in terms of standard deviation compared to several other synthetic datasets. 

\begin{table}[!htb]
    \centering
    \caption{Peformance of models trained on Digi2Real and WebFace dataset}

    \resizebox{0.99\columnwidth}{!}{

    \begin{tabular}{ccrrrrrrr}
        \toprule
        \multicolumn{2}{c}{\textbf{Number of IDs}} & \multicolumn{7}{c}{\textbf{Performance Metrics}} \\
        \cmidrule(lr){1-2} \cmidrule(lr){3-9}
        \textbf{Digi2Real} & \textbf{WebFace} & \textbf{LFW} & \textbf{CALFW} & \textbf{CPLFW} & \textbf{AgeDB30} & \textbf{CFP-FP} & \textbf{IJB-B} & \textbf{IJB-C} \\
        \midrule
20K &0    &  98.58 &  87.22 &  82.65 &    84.72 &   87.49 &  70.14 &  75.80 \\
20K &10   &  98.62 &  87.12 &  82.67 &    84.02 &   87.24 &  67.59 &  73.66 \\
20K &100  &  97.95 &  86.20 &  81.70 &    82.38 &   87.70 &  54.88 &  63.58 \\
20K &200  &  98.53 &  86.48 &  82.47 &    83.88 &   88.56 &  59.89 &  67.54 \\
20K &500  &  98.97 &  89.25 &  84.85 &    88.12 &   89.93 &  72.22 &  78.51 \\
20K &1000 &  99.25 &  91.27 &  86.60 &    89.57 &   91.66 &  76.95 &  82.00 \\
20K &2000 &  99.20 &  91.92 &  87.42 &    91.28 &   92.57 &  81.98 &  86.42 \\
        \bottomrule
    \end{tabular}
    }
    \label{tab:verification_ids_real_synth}
\end{table}

\textbf{Training Models using Synthetic and Real Data} The performance of face recognition models trained solely on synthetic data still falls short of those trained on real data. Although synthetic data can address privacy and legal challenges, the performance gap hinders the practical use of these models in critical applications. Previous studies have shown that incorporating a small amount of real data can help close the performance gap between models trained on synthetic versus real datasets \cite{bae2023digiface}. This suggests a practical approach where consent could be obtained from a limited number of subjects. To explore this, we conducted an ablation study using our synthetic dataset along with samples from WebFace260M, progressively adding identities from WebFace to our synthetic dataset and training new models. The results in Table \ref{tab:verification_ids_real_synth} show significant performance improvements with the addition of even a small number of real identities. Adding 1,000 identities from WebFace, for instance, raises the LFW accuracy to 99.25\%, highlighting the effectiveness of combining real and synthetic data.

\section{Discussions}

In this work, we present a novel approach for enhancing realism in synthetic images generated from a graphics pipeline. Our method introduces intra-class variation through hypersphere interpolation, reducing the gap to real images in CLIP space. We evaluated the performance of the synthesized datasets on standard face recognition benchmarks, demonstrating a significant improvement over the original DigiFace dataset. Additionally, our approach performs comparably to, and often outperforms, many other synthetic datasets. While our method’s performance is limited by the quality of the source dataset, applying it with more advanced graphics pipelines could yield further improvements.

\textbf{Limitations} While this approach successfully generates high-quality images, there are several important aspects to note. Firstly, the Arc2Face training pipeline utilizes images from WebFace42M, FFHQ, and CelebA-HQ, indicating that the Arc2Face model requires supervision from a large number of samples to generate identity-consistent images. Moreover, their method depends on a high-quality face recognition model, also trained on the WebFace dataset, which is a common limitation among many synthetic datasets. Additionally, as shown in the figures, the generated images do not always appear very realistic. 

Despite these limitations, our approach demonstrates that it is possible to train highly effective face recognition models by first procedurally generating samples from a graphics pipeline, followed by generating identity-consistent variations. We hope this inspires researchers to focus more on procedural generation pipelines for synthetic data and to develop more sample-efficient approaches for face recognition.

\section{Conclusion}
In this work, we demonstrate that adding realism to images generated from a graphics pipeline can greatly enhance face recognition performance. The use of the graphics pipeline allows for the creation of a large number of identities with user-defined poses, expressions, accessories, lighting, ethnicities, etc., providing fine-grained control over the generation process. This capability facilitates the easy generation of balanced datasets. The next step is to make the generated images appear photorealistic, and our approach can be employed to achieve this realism. Essentially, our proposed hybrid approach combines the effectiveness of both graphics rendering and data-driven synthetic data generation paradigms, further boosting face recognition performance. We also note the need for more innovation in transferring realism with minimal use of real data. We hope the performance improvement with our approach will encourage researchers to focus more on hybrid approaches to synthetic data generation.

\section*{Acknowledgment}

This research is based on work conducted in the SAFER project and supported by the Hasler Foundation’s Responsible AI program.

\clearpage

%%%%%%%%% REFERENCES
{\small
\bibliographystyle{ieee_fullname}
\bibliography{egbib}
}

\end{document}